\documentclass[runningheads]{llncs}
\usepackage[T1]{fontenc}
\usepackage{graphicx}
\usepackage{amsmath}
\usepackage{amssymb}
\usepackage{booktabs}
\usepackage{multirow}
\usepackage{makecell}
\usepackage{ulem}

\begin{document}

\title{RectiCast: Rectifying Distribution Shift in Cascaded Precipitation Nowcasting}
\titlerunning{Rectifying Distribution Shift in Cascaded Precipitation Nowcasting}

\author{Fanbo Ju \and Haiyuan Shi \and Qingjian Ni\thanks{Corresponding author.}}

\authorrunning{F. Ju et al.}

\institute{Southeast University, Nanjing, China\\
\email{\{jufanbo,haiyuan.shi,nqj\}@seu.edu.cn}}

\maketitle 

\begin{abstract}
Precipitation nowcasting, which aims to provide high spatio-temporal resolution precipitation forecasts by leveraging current radar observations, is a core task in regional weather forecasting. Recently, the cascaded architecture has emerged as the mainstream paradigm for deep learning-based precipitation nowcasting. This paradigm involves a deterministic model to predict posterior mean, followed by a probabilistic model to generate local stochasticity. However, existing methods commonly conflate the systematic distribution shift in deterministic predictions with local stochasticity, which contaminates the probabilistic component and degrades forecast accuracy, especially over longer lead times. To address this issue, we introduce RectiCast, a two-stage framework that explicitly decouples the rectification of mean-field shift from the generation of local stochasticity via a dual Flow Matching model. In the first stage, a deterministic model generates the posterior mean. In the second stage, a Rectifier explicitly learns the distribution shift to yield a rectified mean, conditioning a Generator that models local stochasticity. Experiments on two radar datasets demonstrate that RectiCast achieves significant performance improvements over existing state-of-the-art methods.

\end{abstract}

\section{Introduction}

Precipitation nowcasting, which aims to provide high spatio-temporal resolution forecasts for up to 2 hours based on current observations, is a core task in regional weather forecasting. Conventional deep learning methods are primarily categorized into deterministic and probabilistic models. Deterministic models directly optimize the mean squared error (MSE) against the ground-truth, effectively capturing macroscopic motion trends. However, their predictions lack fine-grained local details. Probabilistic models model the entire precipitation system stochastically. While capable of generating predictions with realistic local details, the predicted location of precipitation is often inaccurate~\cite{diffcast}. Recently, to integrate the strengths of both approaches, several works~\cite{diffcast,percpcast,cascast} have employed a cascaded architecture. This framework treats the deterministic prediction as the posterior mean (i.e., the macroscopic trend) given the observation sequence~\cite{diffcast,percpcast}. A probabilistic model then generates the residual between the ground-truth distribution and this mean (i.e., the local stochasticity)~\cite{diffcast}. This paradigm has proven effective in capturing the evolution of weather systems at multiple scales.

\begin{figure}[t]
\centering
\includegraphics[width=\textwidth]{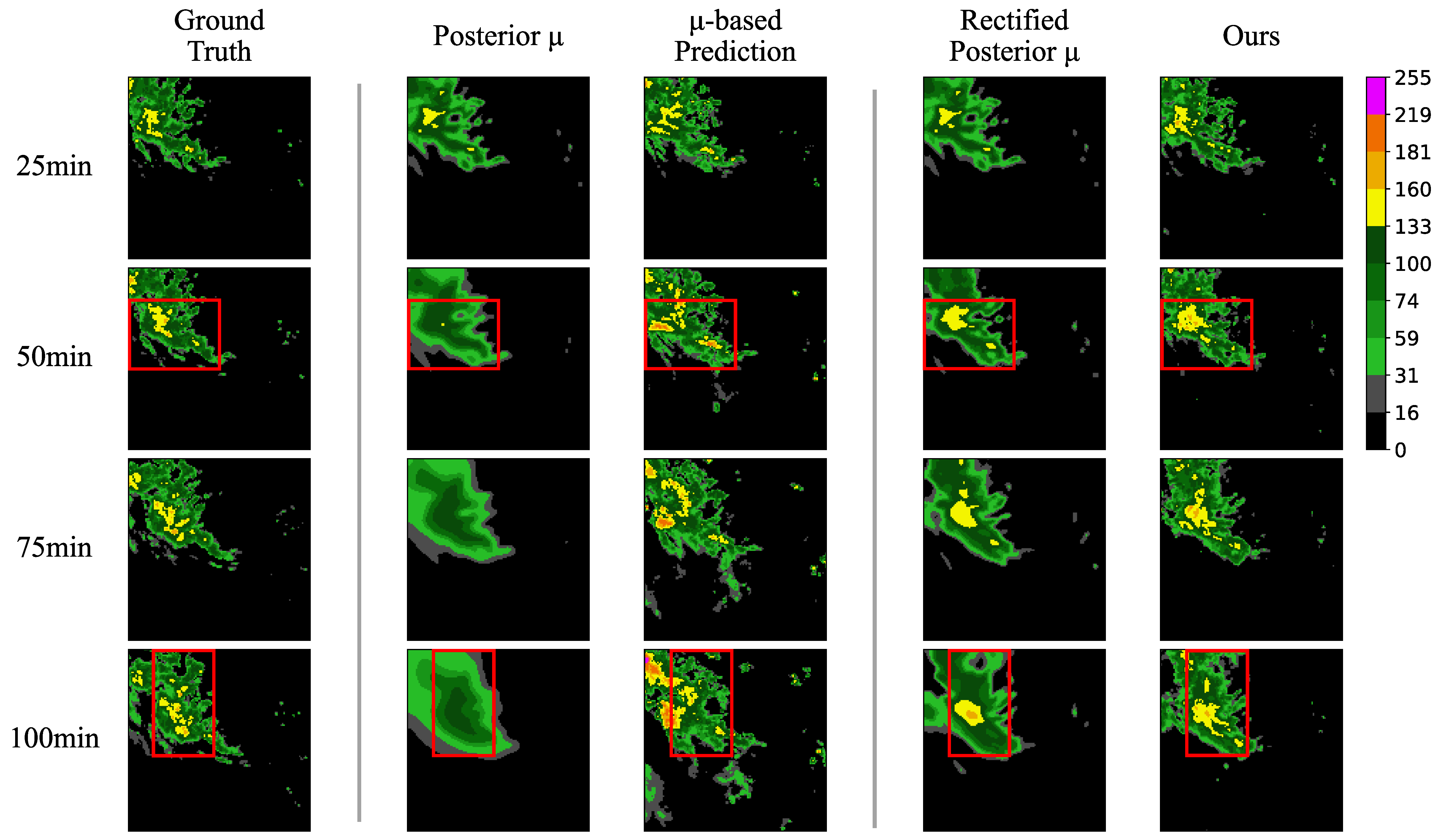}
\caption{Illustration of the distribution shift in deterministic predictions and its impact on cascaded models. The deterministic forecast (Posterior $\mu$) suffers from temporal distribution shift, leading to progressively blurry prediction. Consequently, the probabilistic component conditioned on raw $\mu$ ($\mu$-based Prediction) fails to accurately capture local precipitation morphology and intensity, especially in the red boxes. In contrast, our Rectified Posterior $\mu$ corrects this shift, enabling our final forecast to deliver substantially more accurate intensity and patterns in the highlighted areas.} 
\label{fig:illustration}
\end{figure}

However, existing methods typically treat the raw output of a deterministic model as the posterior mean, overlooking its systematic bias. Deterministic models typically predict the conditional expectation of future precipitation conditioned on past observations, resulting in increasingly ambiguous forecasts as lead times extend and predictive uncertainty accumulates. This mechanism causes a distribution shift in the posterior mean sequence over time, which manifests as forecasts becoming progressively blurrier and high-value echoes fading away. Consequently, the probabilistic component must learn a mixed distribution of the systematic distribution shift and local stochasticity. This causes mutual interference between modeling these two effects, leading to inaccurate forecasts of local precipitation patterns and intensity as illustrated in Figure 1.

To this end, we propose RectiCast, a novel two-stage precipitation nowcasting framework. The framework first generates a posterior mean sequence using a deterministic model. Subsequently, it introduces two synergistic Flow Matching models to explicitly decouple mean-shift rectification from local stochastic generation. The first model, the Rectifier, explicitly learns the temporal distribution shift of the posterior mean sequence to produce a rectified mean sequence. The second model, the Generator, is conditioned on the rectified mean sequence and focuses on learning the local stochasticity of the ground-truth precipitation. Our main contributions are as follows:
\begin{itemize}
    \item We first identify that in existing cascaded nowcasting frameworks, the learning objective of the probabilistic component is biased by the distribution shift of the posterior mean.
    \item We propose RectiCast, a novel framework that decouples mean-shift rectification and local stochastic generation by assigning each task to a dedicated Flow Matching model, and validate its effectiveness.
    \item Experiments on two radar datasets demonstrate that our method achieves significant performance improvements over existing state-of-the-art methods.
\end{itemize}

\section{Related Work}
\label{sec:related}
Deep learning-based precipitation nowcasting methods are generally categorized into deterministic, probabilistic, and cascaded approaches. Deterministic models excel at capturing the overall trend of precipitation movement. ConvLSTM~\cite{convlstm} combines convolutional operators with recurrent units to extract spatio-temporal features, leading to variants such as TrajGRU~\cite{trajgru} and PredRNN~\cite{predrnn}. PhyDNet~\cite{phydnet} decomposes the prediction into PDE-guided physical motion and random motion to enhance physical consistency. SimVP~\cite{simvp} employs a simplified convolutional encoder-decoder architecture for greater computational efficiency. Earthformer~\cite{earthformer} and Earthfarseer~\cite{earthfarseer} introduce Transformer~\cite{transformer} to precipitation nowcasting. However, deterministic models tend to predict an average of multiple future scenarios, leading to forecasts that become blurry over time. Probabilistic models generate more realistic details by modeling the data distribution of future precipitation, but they often fail to accurately capture large-scale precipitation motion. DGMR~\cite{dgmr} introduces spatial and temporal discriminators to constrain the prediction distribution through adversarial training. NowcastNet~\cite{nowcastnet} introduces physical constraints to regularize the forecast of generative backbone. PreDiff~\cite{prediff} constructs a diffusion model in the latent space and designs a guidance module to adjust the denoising process. MCVD~\cite{mcvd} learns the inter-frame relationships of precipitation through random masking. Cascaded models combine the strengths of the previous two approaches, significantly improving forecast accuracy. DiffCast~\cite{diffcast} introduces a residual diffusion mechanism to model local stochastic variations conditioned on a deterministic forecast. CasCast~\cite{cascast} performs the diffusion process in latent space, generating the entire forecast sequence at once. PercpCast~\cite{percpcast} employs a perceptual loss to constrain the generation process, achieving strong results in long-range forecasting.

\section{Problem Setup and Preliminaries}
\label{sec:problem}

\subsection{Problem Setup}
Precipitation nowcasting is formulated as a spatio-temporal prediction problem. Given the current radar observations $x_{0:T} \in \mathbb{R}^{T \times C \times H \times W}$, the objective of the nowcasting model is to predict the radar echos $y_{T:T+T'} \in \mathbb{R}^{T' \times C \times H \times W}$ for the future $T'$ frames. $H$ and $W$ represent the spatial dimensions of each frame. The channel dimension $C$ is set to 1 since radar echo data is single-channel.

\subsection{Preliminary: Flow Matching}
Flow Matching~\cite{flow} is an emerging generative framework. Its core idea is to learn the conditional velocity along a linear path from pure Gaussian noise to a real data sample.
Let $f_{\theta}$ denote the generative model, $\epsilon \sim \mathcal{N}(0,I)$ be Gaussian noise, $x$ be a real data sample from the training set, and $t \sim U(0,1)$ be a timestep uniformly sampled from $[0,1]$. The intermediate state $x_t$ on this linear path is defined as:
\begin{equation}
x_t = (1-t) \epsilon + tx
\label{eq:fm_path}
\end{equation}
The training objective for the model is to approximate the  velocity $v = x - \epsilon$. The loss function is:
\begin{equation}
L_{FM} = \|f_{\theta}(x_t, t) - (x-\epsilon) \|^2
\label{eq:fm_loss}
\end{equation}
Here, the velocity $v = x-\epsilon$ is the derivative of $x_t$ with respect to $t$. The model takes $x_t$ and $t$ as input, where $t$ represents the noise level. During inference, Flow Matching generates samples by solving the ordinary differential equation (ODE) defined by the predicted velocity field $f_{\theta}(x_t, t)$. Starting from Gaussian noise $x_0 = \epsilon$, the trajectory is approximated numerically over discrete timesteps, typically via the Euler method:
\begin{equation}
x_{t+\Delta t} = x_t + \Delta t \cdot f_{\theta}(x_t, t)
\label{eq:fm_euler}
\end{equation}
Benefiting from the straight trajectory in Eq.~\ref{eq:fm_path}, Flow Matching allows larger time steps $\Delta t$ compared to diffusion models. This property allows for high-quality sampling with fewer function evaluations, thus accelerating inference.

\section{Methodology}
\label{sec:method}
\begin{figure}[t]
\centering
\includegraphics[width=\textwidth]{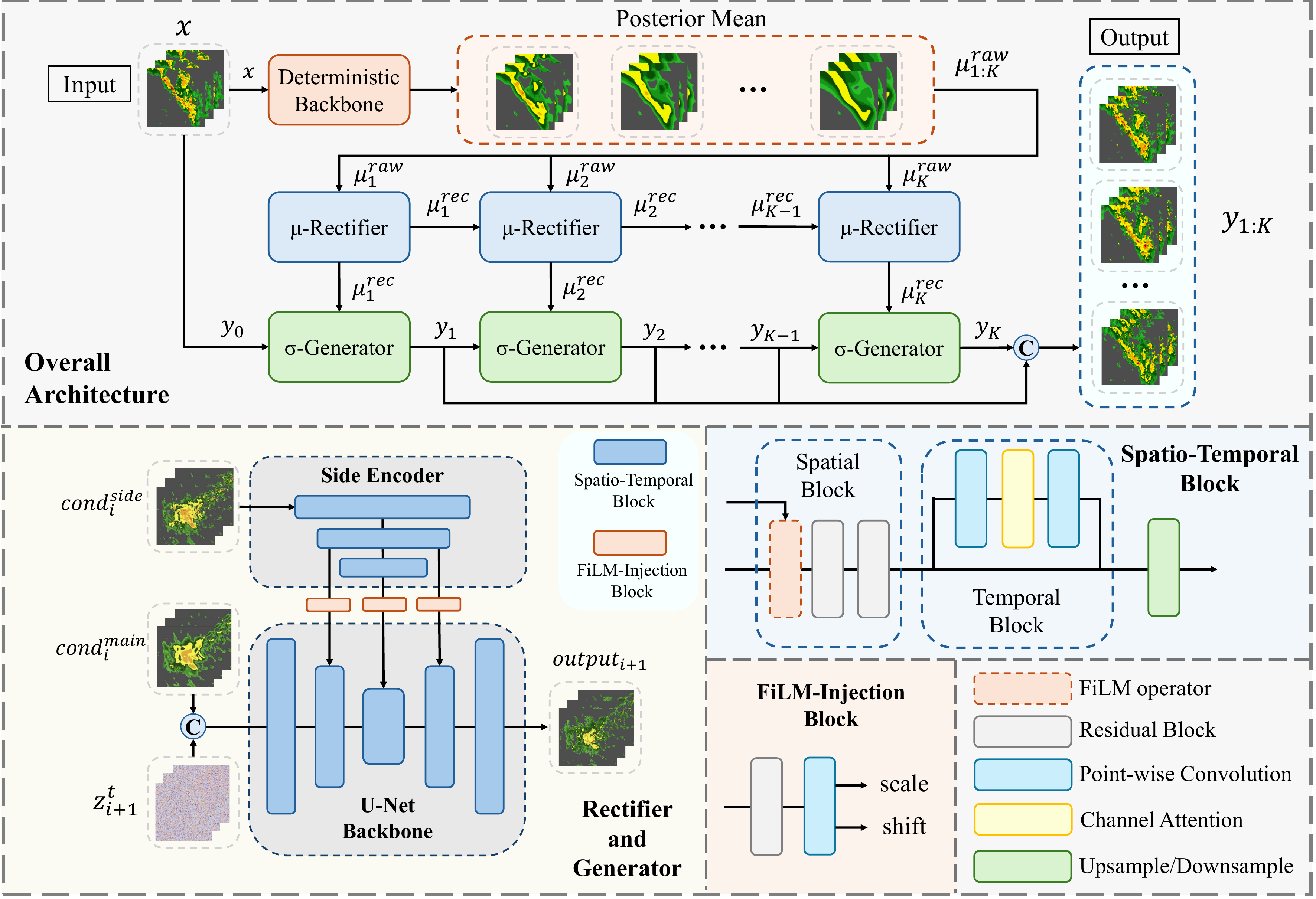}
\caption{The overall architecture of our proposed framework.} \label{fig:architecture}
\end{figure}

Our proposed method consists of two main components: a Deterministic Predictive Backbone and a Rectifier-Generator Framework. The input sequence is first processed by the Deterministic Predictive Backbone to generate an initial forecast. This forecast is then refined by the Rectifier-Generator Framework. The overall architecture is illustrated in Fig.~\ref{fig:architecture}.

\subsection{Deterministic Predictive Backbone}
\label{ssec:backbone}
In the first stage of prediction, we use a deterministic predictive backbone $D_{\theta_1}$ to predict the posterior mean $\mu^{\text{raw}}$. Similar to~\cite{diffcast}, our framework supports any prediction models of recurrent-based and recurrent-free architectures. We train the deterministic predictive backbone to minimize the mean-squared error (MSE) between the posterior mean and the ground truth $y$:
\begin{equation}
\mathcal{L}_{\theta_1} = \mathbb{E}(\Vert\mu^{\text{raw}} - y\Vert^2)
\label{eq:mse}
\end{equation}
This enables $D_{\theta_1}$ to model the distribution of $p_{\theta_1}(\mu^{\text{raw}}|x)$, which is the distribution of the posterior mean of $y$ given the current observation $x$. The prediction $\mu^{\text{raw}}$ obtained this way effectively captures the global motion trend and serves as a strong denoising prior for the subsequent probabilistic component.

\subsection{Rectifier-Generator Framework}
\label{ssec:rectifier-generator}
As mentioned, conditioning the probabilistic model directly on $\mu^{\text{raw}}$ conflates its learning objectives. To explicitly disentangle the distribution shift from the local stochasticity, we train two separate Flow Matching models in the second stage to model each component.

The first model, the Rectifier $\mathcal{R}_{\theta_2}$, models the distribution shift of the raw posterior mean sequence $\mu^{\text{raw}}$ over lead time. While a posterior mean entirely free of distribution shift is unattainable, the shift in predictions with shorter lead times remains within an acceptable range for modeling local stochasticity. Consequently, we utilize the initial segment of $\mu^{\text{raw}}$ as the target distribution for $\mathcal{R}_{\theta_2}$. Following the optimal configuration in~\cite{diffcast}, the segment length $m$ is set equal to the input sequence length. We model the rectification process autoregressively via a Spatial-Temporal U-Net equipped with a side encoder, which will be discussed in 4.3. We define the target for the $i$-th segment as $\mu^{\text{rec}}_i = D_{\theta_1}(s_{i-1})_1$, for $i=2, 3, \dots, \lceil L_{\text{out}} / m \rceil$, where $D_{\theta_1}(s_{i-1})_1$ represents the first segment of the backbone prediction given the previous ground truth segment $s_{i-1}$. For the first segment ($i=1$), no rectification applies, i.e., $\mu^{\text{rec}}_{1} = \mu^{\text{raw}}_{1}$. The Rectifier is conditioned on the current raw mean segment $\mu^{\text{raw}}_{i}$ and the previous rectified mean segment $\mu^{\text{rec}}_{i-1}$. In addition to the timestep $t$, the network takes the segment index $i$ as a condition to indicate the degree of shift. For the $i$-th segment, $\mathcal{R}_{\theta_2}$ establishes the following conditional distribution:
\begin{equation}
    p_{\theta_2} (\mu^{\text{rec}}_i | \mu^{\text{raw}}_{i}, \mu^{\text{rec}}_{i-1}, i)
    \label{eq:rectifier_dist}
\end{equation}
$\mathcal{R}_{\theta_2}$ is trained by minimizing the following loss:
\begin{equation}
    \mathcal{L}_{\theta_2} = \mathbb{E}_{\mu^{\text{rec}}_i, \mu^{\text{raw}}_{i}, \mu^{\text{rec}}_{i-1}, i, t, \epsilon} \left( \left\Vert \mathcal{R}_{\theta_2}(z_t, \mu^{\text{raw}}_{i}, \mu^{\text{rec}}_{i-1}, i, t) - (\mu^{\text{rec}}_i - \epsilon) \right\Vert^2 \right)
    \label{eq:rectifier_loss}
\end{equation}
where $z_t = (1 - t) \epsilon + t \mu^{\text{rec}}_i$, and $\epsilon \sim \mathcal{N}(0,I)$.

The second model, the Generator $\mathcal{G}_{\theta_3}$, focuses on modeling local stochasticity. It employs the same network architecture as $\mathcal{R}_{\theta_2}$ and forecasts autoregressively. $\mathcal{G}_{\theta_3}$ is conditioned on the prediction from the previous segment $\hat{y}_{i-1}$ and the current rectified mean $\mu^{\text{rec}}_{i}$. For the $i$-th segment, $\mathcal{G}_{\theta_3}$ models:
\begin{equation}
    p_{\theta_3}(y_i | \mu^{\text{rec}}_{i}, \hat{y}_{i-1})
    \label{eq:generator_dist}
\end{equation}
where $i=1, 2, \dots, \lceil L_{\text{out}} / m \rceil$, and $\hat{y}_i$ is the Generator prediction with $\hat{y}_0 = x$. $\mathcal{G}_{\theta_3}$ is trained by minimizing:
\begin{equation}
    \mathcal{L}_{\theta_3} = \mathbb{E}_{y_i, \hat{y}_{i-1}, \mu^{\text{rec}}_{i}, t, \epsilon} \left( \left\Vert \mathcal{G}_{\theta_3}(z_t, \hat{y}_{i-1}, \mu^{\text{rec}}_{i}, t) - (y_i - \epsilon) \right\Vert^2 \right)
    \label{eq:generator_loss}
\end{equation}
where $z_t = (1 - t) \epsilon + t y_i$, and $\epsilon \sim \mathcal{N}(0,I)$.

\subsection{Network Architecture}
This subsection details the architecture of the Rectifier and Generator. The architectures of both the Rectifier and Generator consist of a U-Net backbone and a side encoder. The structure of the side encoder is similar to that of the U-Net encoder. Each Spatio-Temporal Block in the U-Net and the side encoder is composed of several spatial residual blocks and a temporal attention block. The spatial residual blocks independently extract spatial features from each input frame. The temporal attention block employs bottleneck point-wise convolutions and channel attention mechanisms to model inter-frame dependencies. Conditions for Flow Matching are injected through both the backbone and the side encoder. The backbone condition is concatenated with $z_t$ frame-by-frame along the channel dimension. The side-injected condition is fed into the side encoder to generate scale and shift parameters, which are then injected via FiLM~\cite{film} before the spatial residual blocks at the corresponding scales in the U-Net. To prevent the model from learning shortcuts, the conditioning from the side network is injected only into deeper layers of the U-Net backbone.

For the Rectifier, the raw mean of the current segment $\mu_i^{\text{raw}}$ is conditioned through the backbone network, while the rectified mean from the previous segment $\mu_{i-1}^{\text{rec}}$ is injected via the side encoder. This design encourages the model to focus on spatial dependencies while generating a more temporally consistent sequence of rectified means. For the Generator, the prediction from the previous segment $\hat{y}_{i-1}$ contains rich high-frequency details, while the rectified mean $\mu_i^{\text{rec}}$ is better for guiding large-scale features. Therefore, $\hat{y}_{i-1}$ is fed into the backbone, while $\mu_i^{\text{rec}}$ is injected through the side encoder.

\subsection{Two-Staged Training}
We employ a two-stage training strategy. In the first stage, we train the deterministic backbone. In the second stage, we freeze the deterministic backbone and train the Rectifier and Generator in parallel. During the second stage, we adopt a teacher forcing strategy. Specifically, the Rectifier's input $\mu_{i-1}^{\text{rec}}$ is replaced with the ground-truth rectified mean from the previous segment $D_{\theta_1}(s_{i-2})$. The Generator's inputs $\mu_i^{\text{rec}}$ and $\hat{y}_{i-1}$ are replaced with $D_{\theta_1}(s_{i-1})$ and $s_{i-1}$, respectively. This two-stage approach allows the Rectifier and Generator to learn from a well-trained deterministic backbone, thus enhancing the training stability.

\section{Experiment}
\label{sec:exp}

\subsection{Experimental Setting}
\textbf{Dataset:} Experiments are conducted on SEVIR~\cite{sevir} and MeteoNet~\cite{meteonet}. SEVIR contains 20,394 sequences of radar frames for weather events from 2017 to 2020. Each radar sequence covers a 384~km~$\times$~384~km area with a 1~km resolution, comprising 49 frames captured every 5 minutes over a 4-hour period. We split the dataset into training, validation, and test sets with the time points January 1, 2019, and July 1, 2019, respectively. Samples of 25 consecutive frames are extracted from each sequence with a stride of 5 frames. MeteoNet contains precipitation radar data from northwestern and southeastern France from 2016 to 2018. The radar sequences cover a 550~km~$\times$~550~km area with a 5-minute observation interval. Dataset splitting and anomaly filtering of MeteoNet follow~\cite{diffcast}. The spatial dimensions of all datasets are downsampled to 128$\times$128. The prediction task is to forecast the next 20 frames given 5 initial frames.

\smallskip 
\noindent \textbf{Evaluation:} We evaluate the model's nowcasting performance using the average Critical Success Index (CSI) and Heidke Skill Score (HSS), calculated as in~\cite{prediff,recon,sevir}. CSI measures pixel-wise matching between predictions and ground truth over a set of thresholds, while HSS measures the forecast skill improvement relative to random forecast. We follow~\cite{diffcast} for data scaling and threshold selection on both SEVIR and MeteoNet datasets. To evaluate forecast accuracy at different spatial scales, we also compute CSI on predictions downsampled via 4$\times$4 and 16$\times$16 max-pooling following~\cite{diffcast}. Additionally, we calculate the Structural Similarity (SSIM) and the Learned Perceptual Image Patch Similarity (LPIPS) to evaluate the visual quality of the forecasts.

\smallskip
\noindent \textbf{Baselines:} Seven notable models are selected as baselines, including three deterministic models, one probabilistic model, and three cascaded models. ConvLSTM~\cite{convlstm}, SimVP~\cite{simvp}, and Earthformer~\cite{earthformer} are deterministic models based on RNN, CNN, and Transformer, respectively. Prediff~\cite{prediff} is a probabilistic model based on the Latent Diffusion Model. DiffCast~\cite{diffcast}, CasCast~\cite{cascast}, and PercpCast~\cite{percpcast} are recent state-of-the-art cascaded models.

\smallskip
\noindent \textbf{Implementation Details:} All models were trained using the Adam optimizer with a cosine learning rate scheduler. The maximum and minimum learning rates were $10^{-4}$ and $10^{-7}$, respectively. We trained the models for 200k iterations on SEVIR and 80k iterations on MeteoNet. All cascaded models were trained in two stages. To ensure a fair comparison of the probabilistic components, SimVP~\cite{simvp} was used uniformly as the deterministic component in all cascaded models. During inference, diffusion-based and flow-based modules employ the DDIM sampler~\cite{ddim} and the Euler solver, respectively. The number of sampling steps following the experimental settings in their papers. All experiments were conducted on 4 V100 GPUs with a global batch size of 24.

\begin{table}[t]
\centering
\caption{Quantitative comparison on the SEVIR and MeteoNet datasets. Best and second-best results are marked in \textbf{bold} and \uline{underlined}, respectively. For all metrics except LPIPS, higher is better ($\uparrow$). For LPIPS, lower is better ($\downarrow$).}\label{tab:results_sevir_meteonet}
\resizebox{\textwidth}{!}{%
\begin{tabular}{l|cccccc|cccccc}
\toprule
\multirow{2}{*}{\textbf{Model}} & \multicolumn{6}{c|}{\textbf{SEVIR}} & \multicolumn{6}{c}{\textbf{MeteoNet}} \\
\cmidrule{2-13}
& CSI$\uparrow$ & CSI4$\uparrow$ & CSI16$\uparrow$ & HSS$\uparrow$ & SSIM$\uparrow$ & LPIPS$\downarrow$ & CSI$\uparrow$ & CSI4$\uparrow$ & CSI16$\uparrow$ & HSS$\uparrow$ & SSIM$\uparrow$ & LPIPS$\downarrow$ \\
\midrule
SimVP & 0.2873 & 0.3014 & 0.3335 & 0.3543 & 0.6291 & 0.3909 & 0.3382 & 0.3391 & 0.3980 & 0.4589 & 0.7394 & 0.3622 \\
ConvLSTM & 0.2792 & 0.2903 & 0.3268 & 0.3412 & 0.6067 & 0.3928 & 0.3394 & 0.3555 & 0.4021 & 0.4633 & 0.7756 & 0.3001 \\
Earthformer & 0.2519 & 0.2704 & 0.3046 & 0.3016 & \textbf{0.6521} & 0.4112 & 0.3244 & 0.3403 & 0.4193 & 0.4554 & 0.7691 & 0.3482 \\
Prediff & 0.2377 & 0.3124 & 0.4211 & 0.3096 & 0.5423 & 0.2804 & 0.2479 & 0.3692 & 0.5377 & 0.3523 & 0.6703 & 0.2077 \\
\midrule
DiffCast & \uline{0.3105} & \uline{0.3988} & \uline{0.5641} & 0.4028 & 0.6305 & \textbf{0.1836} & 0.3492 & \uline{0.4734} & \uline{0.6407} & \uline{0.4783} & \uline{0.7859} & \textbf{0.1204} \\
PercpCast & 0.3103 & 0.3946 & 0.5239 & \uline{0.4052} & \uline{0.6429} & 0.1997 & \uline{0.3532} & 0.4629 & 0.6018 & 0.4721 & 0.7828 & 0.1367 \\
CasCast & 0.3041 & 0.3836 & 0.5432 & 0.3889 & 0.6129 & 0.1977 & 0.3327 & 0.4439 & 0.6188 & 0.4511 & 0.7807 & 0.1256 \\
\midrule
\textbf{RectiCast} & \textbf{0.3269} & \textbf{0.4139} & \textbf{0.5806} & \textbf{0.4201} & 0.6331 & \uline{0.1894} & \textbf{0.4001} & \textbf{0.5049} & \textbf{0.6684} & \textbf{0.5339} & \textbf{0.7899} & \uline{0.1211} \\
\bottomrule
\end{tabular}%
}
\end{table}

\subsection{Experimental Results}

\begin{figure}[t]
\centering
\includegraphics[width=\textwidth]{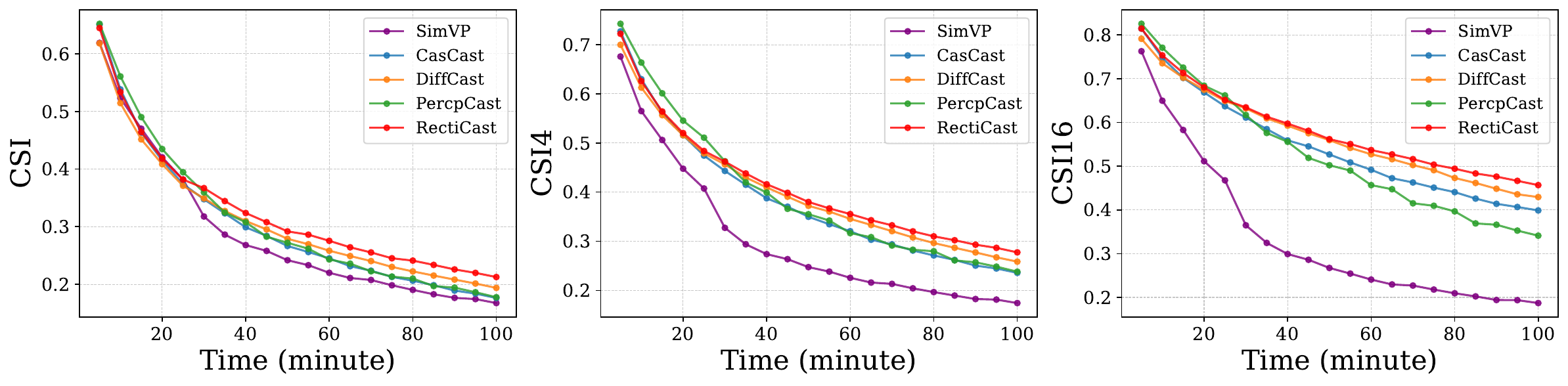}
\caption{CSI curves against lead time for cascaded models on the SEVIR dataset.}\label{fig:csi_curves}
\end{figure}

Table~\ref{tab:results_sevir_meteonet} presents the experimental results on the SEVIR and MeteoNet datasets. From these results, we make the following observations: (i) RectiCast retains the strengths of cascaded models, outperforming both deterministic and probabilistic baselines on most metrics. (ii) Compared to all cascaded baselines, RectiCast achieves 5\% to 13\% improvements on pixel-wise CSI and HSS, demonstrating its superior capability in generating small-scale precipitation. This indicates that RectiCast's decoupled learning strategy enables the probabilistic component to learn the local stochasticity of precipitation more accurately. (iii) RectiCast shows improvements on CSI4 and CSI16 over all cascaded baselines, suggesting that decoupling the learning of distribution shift of deterministic forecast effectively mitigates its negative impact on the accurate forecasting of larger-scale patterns. (iv) RectiCast maintains LPIPS and SSIM scores comparable to other cascaded models, indicating that its forecasts are consistent with the true precipitation distribution.

To investigate how the decoupled learning architecture improves the forecasting capability of RectiCast, we plot the three CSI curves of all cascaded models and their deterministic component SimVP~\cite{simvp} against lead time in Figure~\ref{fig:csi_curves}. As the lead time increases, prediction uncertainty grows, causing the performance of all models to decline. However, the performance of different cascaded models declines at different rates. This suggests that the cascaded structures are affected differently by the distribution shift of the deterministic prediction. PercpCast~\cite{percpcast} performs best at short lead times, but its performance drops sharply as SimVP's performance declines. This is because its probabilistic part directly models a frame-wise mapping from the deterministic prediction to the final forecast, making it the most susceptible to the distribution shift of the deterministic prediction. While DiffCast~\cite{diffcast} and CasCast~\cite{cascast} do not explicitly model the distribution shift of the deterministic prediction, their modeling of inter-frame dependencies allows them to maintain better performance at longer lead times. RectiCast's three CSI curves decline the most slowly over time, suggesting that its explicit modeling of the deterministic prediction's distribution shift significantly enhances its long-term forecasting capability. This again validates the effectiveness of RectiCast's Rectifier-Generator framework.

\begin{figure}[t]
\centering
\includegraphics[width=\textwidth]{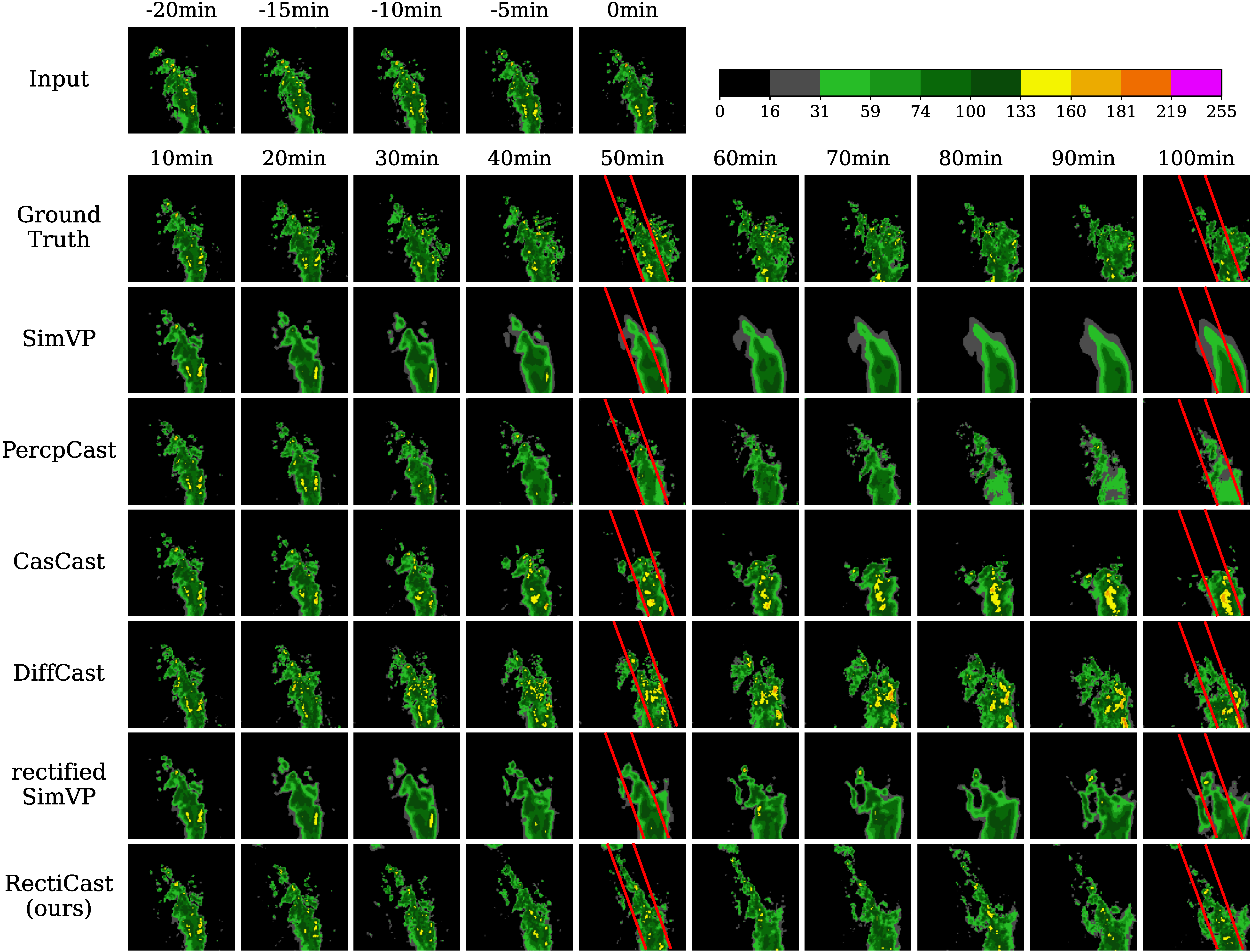}
\caption{A case study from different cascaded models on the SEVIR dataset.}\label{fig:case_study}
\end{figure}

Figure~\ref{fig:case_study} presents a case study on the SEVIR dataset, comparing the results of all cascaded models and their deterministic component SimVP. Although SimVP's prediction accurately captures the overall motion trend, it exhibits increasing blurriness and detail degradation over time. Consequently, when modeling the main rainband highlighted by the red lines, PercpCast, DiffCast, and CasCast all show degraded morphology and inaccurate intensity with increasing lead time. The rectified SimVP sequence, corrected by RectiCast’s Rectifier, partially mitigates this degradation and preserves correct morphological features within the highlighted region. As a result, the final forecast of RectiCast is more consistent with the ground truth in both precipitation intensity and rainband morphology. These observations again affirm the superiority of our proposed framework.

\subsection{Ablation Study}

\begin{table}[t]
\centering
\caption{Ablation study on the SEVIR dataset. Best results are in \textbf{bold}.}\label{tab:ablation}
\begin{tabular}{l|cccccc}
\toprule
\textbf{Configuration} & CSI$\uparrow$ & CSI4$\uparrow$ & CSI16$\uparrow$ & HSS$\uparrow$ & SSIM$\uparrow$ & LPIPS$\downarrow$ \\
\midrule
Backbone-only  & 0.2873 & 0.3014 & 0.3335 & 0.3543 & 0.6291 & 0.3909 \\
\hline
w/o Rectifier (learns $y$) & 0.3071 & 0.3901 & 0.5598 & 0.3957 & 0.6180 & 0.2201 \\
w/o Rectifier (learns $y - \mu^{\text{raw}}$) & 0.3161 & 0.3907 & 0.5492 & 0.4080 & 0.6330 & 0.2143 \\
\hline
w/o Generator & 0.3016 & 0.3882 & 0.5148 & 0.4033 & 0.6267 & 0.3024 \\
\hline
Full Model & \textbf{0.3269} & \textbf{0.4139} & \textbf{0.5806} & \textbf{0.4201} & \textbf{0.6331} & \textbf{0.1894} \\
\bottomrule
\end{tabular}
\end{table}

\begin{table}[t]
\centering
\caption{Single-sample inference efficiency comparison of all cascaded models.}
\label{tab:efficiency}
\setlength{\tabcolsep}{4pt}
\begin{tabular}{l|ccc|c}
\toprule
\textbf{Metric} & \textbf{DiffCast} & \textbf{CasCast} & \textbf{PercpCast} & \textbf{RectiCast (Ours)} \\
\midrule
NFE & $ 800 $ & $ 400 $ & $ 20 $ & $ 160 $ \\
Runtime (s) & 6.86 & 2.67 & 2.24 & 3.43 \\
\bottomrule
\end{tabular}
\end{table}

To verify the validity of the individual components in RectiCast, we conducted ablation experiments on the SEVIR dataset. We compared the nowcasting performance across several configurations: the full model with both the Rectifier and Generator, a variant without the Generator, and variants without the Rectifier. Without the Generator, the model degenerates into a pure mean rectifier. When the Rectifier is removed, we report metrics for two scenarios: the Generator learning the precipitation distribution $y$ directly, and learning the residual between the distribution and the raw mean $y - \mu^{\text{raw}}$. As shown in Table~\ref{tab:ablation}, ablating either the Rectifier or the Generator leads to a significant performance drop. This indicates that without the Generator, the model fails to generate precise local details. Moreover, the absence of the Rectifier causes the mean shift to interfere with the learning of local stochastics. These results confirm the effectiveness of decoupling the learning of the distribution shift and local stochastics.

Finally, we analyze the computational efficiency in Table~\ref{tab:efficiency}. Although RectiCast's autoregressive manner and dual Flow Matching architecture increases the Number of Function Evaluations (NFE), it maintains fast inference speed, validating a favorable accuracy-efficiency trade-off.

\section{Conclusion}
In this paper, we propose RectiCast to address the distribution shift of the deterministic backbone in cascaded precipitation nowcasting methods. By introducing a Rectifier to explicitly rectify the distribution shift and a Generator to focus on modeling local stochasticity, we effectively decouple these two conflicting objectives. Experiments on SEVIR and MeteoNet datasets demonstrate that RectiCast significantly outperforms existing state-of-the-art methods, particularly over longer lead times. Future work will focus on integrating various atmospheric variables to develop a unified precipitation forecasting system.

\bibliographystyle{splncs04}
\bibliography{ref}

\end{document}